\documentclass[10pt, journal,twoside]{IEEEtran}

\usepackage{times}
\usepackage{epsfig}
\usepackage{graphicx}
\usepackage{amsmath}
\usepackage{amssymb}
\usepackage{algorithm}
\usepackage{algorithmic}
\usepackage{color}

\usepackage{threeparttable}
\usepackage{multirow}
\usepackage{booktabs}
\usepackage{array}
\usepackage{setspace}
\usepackage{fancyhdr}

\usepackage{xcolor}
\usepackage{tikz,xcolor}

\usepackage[backref]{hyperref} 

\hypersetup{hidelinks}
\definecolor{lime}{HTML}{A6CE39}
\DeclareRobustCommand{\orcidicon}{%
    \begin{tikzpicture}
    \draw[lime, fill=lime] (0,0)
    circle [radius=0.16]
    node[white] {{\fontfamily{qag}\selectfont \tiny ID}};    \draw[white, fill=white] (-0.0625,0.095)
    circle [radius=0.007];    \end{tikzpicture}
    \hspace{-2mm}}
\foreach \x in {A, ..., Z}{%
    \expandafter\xdef\csname orcid\x\endcsname{\noexpand\href{https://orcid.org/\csname orcidauthor\x\endcsname}{\noexpand\orcidicon}}
    }

\usepackage{xspace}
\newcommand{\etal}{{\emph{et al.}}}
\newcommand{\eg}{{\emph{e.g.}},\xspace}
\newcommand{\ie}{{\emph{i.e.}},\xspace}

\usepackage[switch]{lineno}

\ifCLASSINFOpdf

\else

\fi

\begin{document}

\title{To be Critical: Self-Calibrated Weakly Supervised Learning for Salient Object Detection}

\author{Yongri~Piao\orcidB{}, Jian~Wang, Miao~Zhang\orcidC{}, Zhengxuan~Ma,
        and Huchuan Lu\orcidA{},~\IEEEmembership{Senior Member,~IEEE}\vspace{-2em}
\thanks{Manuscript received xxx; revised xxx; accepted xxx. 

The associate editor coording the review of this manuscript and approving it for publication was C. L. Philip Chen. (Corresponding author: Miao Zhang)}
\thanks{Yongri Piao, Jian Wang and Huchuan Lu are with the School of Information and Communication Engineering, Dalian University of Technology, China (e-mail: yrpiao@dlut.edu.cn; dlyimi@mail.dlut.edu.cn; lhchuan@dlut.edu.cn).}
\thanks{Miao Zhang and Zhengxuan Ma is with the DUT-RU International School of Information Science \& Engineering, Dalian University of Technology and Key Laboratory for Ubiquitous Network and Service Software of Liaoning Province, China (e-mail: miaozhang@dlut.edu.cn, mzx\_dlut@mail.dlut.edu.cn).}}

\markboth{IEEE TRANSACTIONS ON IMAGE PROCESSING}
{Piao \MakeLowercase{\textit{et al.}}: To be Critical: Self-Calibrated Weakly Supervised Learning for Salient Object Detection}

\maketitle

\begin{abstract}
Weakly-supervised salient object detection (WSOD\footnote{In this paper, we denote weakly supervised salient object detection methods using image-level labels as WSOD for convenience.}) aims to develop saliency models using image-level annotations. Despite of the success of previous works, explorations on an effective training strategy for the saliency network and accurate matches between image-level annotations and salient objects are still inadequate. In this work, 
\textbf{1)} we propose a self-calibrated training strategy by explicitly establishing a mutual calibration loop between pseudo labels and network predictions, liberating the saliency network from error-prone propagation caused by pseudo labels.
\textbf{2)} we prove that even a much smaller dataset (merely $1.8$\% of ImageNet) with well-matched annotations can facilitate models to achieve better performance as well as generalizability. This sheds new light on the development of WSOD and encourages more contributions to the community. 
Comprehensive experiments demonstrate that our method outperforms all the existing WSOD methods by adopting the self-calibrated strategy only. Steady improvements are further achieved by training on the proposed dataset. Additionally, our method achieves $94.7$\% of the performance of fully-supervised methods on average.
And what is more, the fully supervised models adopting our predicted results as "ground truths" achieve successful results ($95.6$\% for BASNet and $97.3$\% for ITSD on F-measure), while costing only $0.32$\% of labeling time for pixel-level annotation.
\end{abstract}


\IEEEpeerreviewmaketitle

\section{Introduction}

\IEEEPARstart{S}{alient} object detection (SOD) aims to segment objects in an image that visually attract human attention most. It plays an important role in many computer vision and robotic vision tasks~\cite{2014Salient}, such as image segmentation~\cite{Li_2014_CVPR} and visual tracking~\cite{hong2015online}. Recently, deep learning based methods \cite{liu2016dhsnet, wang2016saliency, feng2019attentive, deng2018r3net, DAFNet, DPANet, crm2019tip, crm2018tip} have proved its superiority and achieved remarkable progress. Success of those methods, however, heavily relies a large number of highly accurate pixel-level annotations, which are time-consuming and labor-intensive to collect. A trade-off between testing accuracy and training annotation cost has long existed in the SOD task.  

To alleviate this predicament, several attempts have been made to explore different weakly supervised formats, such as noisy label~\cite{NIPS2019_8314, siva2013looking}, scribble \cite{zhang2020weakly, Yu2020StructureConsistentWS} and image-level annotation (\ie classification label). Image-level annotation based WSOD methods usually adopt a two-stage scheme, which leverages a classification network to generate pseudo labels and then trains a saliency network on these labels.
In this paper, we focus on this most challenging problem of developing WSOD by only using image-level annotation.

\begin{figure}
\vspace{-0mm}
\begin{center}
\includegraphics[width=1\linewidth]{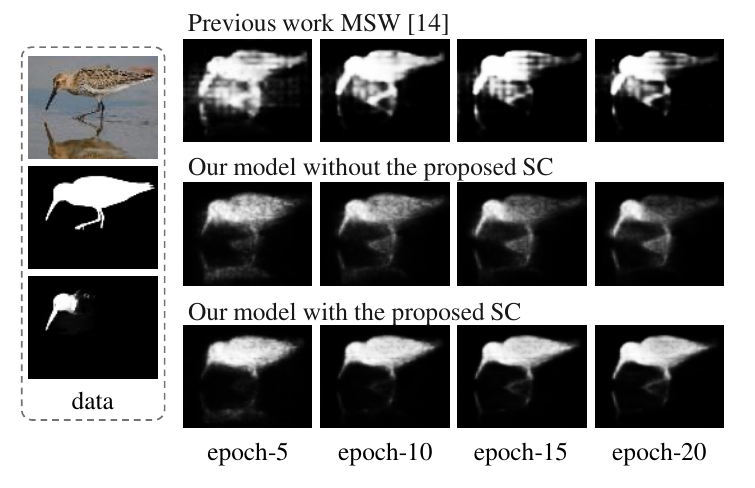}
\end{center}
\vspace{-5mm}
\caption{ The visual saliency predictions during the training process of different models, in which \textbf{SC} represents our proposed self-calibrated training strategy. Column \textbf{data} represents image, ground truth and pseudo label, noting that the ground truth is just for exhibition and not used in our framework.}
\label{introduction2}
\vspace{-2mm}
\end{figure}

Some pioneering works~\cite{wang2017learning, li2018weakly, zeng2019multi} pursue accurate pseudo labels to train a saliency network and achieve good performance. However, given the fact that pseudo labels are still a far cry from the ground truths, the error remaining unaddressed in the pseudo labels can propagate to the generated predictions. This is consistent with the fact that as the number of epochs increases, the parameters of the model are updated and the prediction curve goes from underfitting to optimal to overfitting. 
Interestingly, we observe that the relatively good results containing global representations of saliency can be predicted at the early training process (\eg epoch-5), while the predictions are more prone to error at the latter training process (\eg epoch-20), as shown in the first two rows in Figure {\color{red}\ref{introduction2}}. This inspires us to go one step further exploring how this global representation can be evolved as the model is properly trained.

Moreover, previous works adopt existing large-scale datasets, \eg ImageNet~\cite{imagenet_cvpr09} and COCO~\cite{lin2014microsoft}, to perform WSOD. However, an observable fact should not be ignored that there is an inherent inconsistency between image classification and SOD task. For example, many classification labels do not match the salient objects in both single-object and multi-object cases in ImageNet, as illustrated in Figure {\color{red}\ref{introduction1}}. Such cross-domain inconsistency caused by those mismatched samples impairs the generalizability of models and prevents WSOD methods from achieving optimal results. 

In this work, our core insight is that we can design a self-calibrated training strategy and exploit saliency-based image-level annotations to address the aforementioned challenges. To be specific, we
\textbf{1)} aim to calibrate our network with progressively updated labels to curb the spread of errors in low-quality pseudo labels during the training process.  
\textbf{2)} develop reliable matches for which image-level annotations are correctly corresponding to salient objects.
The source code will be released upon publication.
Concretely, our contributions are as follows:

\begin{figure}
\vspace{-0mm}
\begin{center}
\includegraphics[width=1\linewidth]{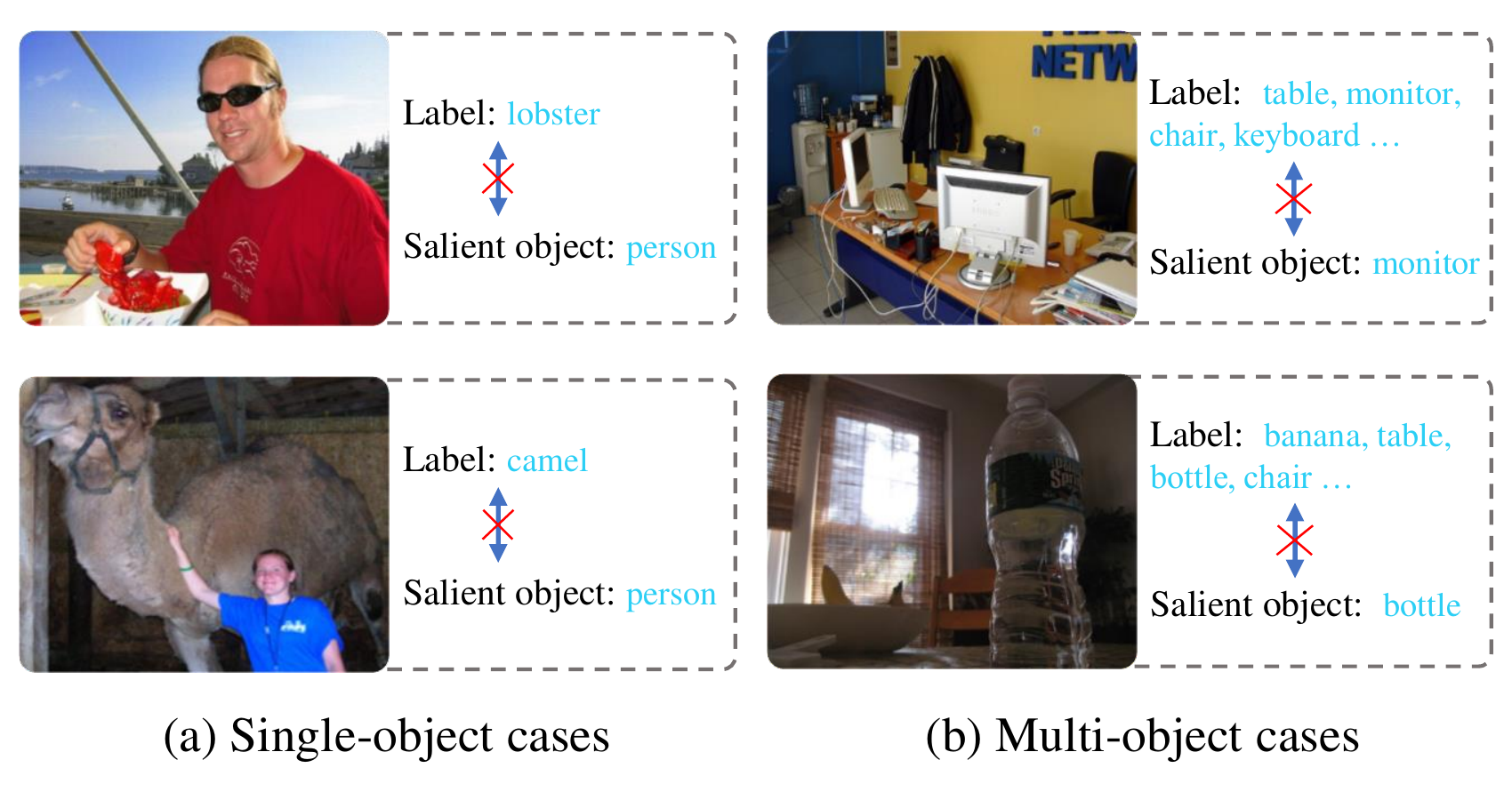}
\end{center}
\vspace{-5mm}
\caption{ Cross-domain inconsistency between ImageNet dataset and salient object detecion. (a) and (b) represent single-object and multi-object cases, respectively.}
\label{introduction1}
\vspace{-2mm}
\end{figure}

\begin{itemize}
\hyphenpenalty=1000
\tolerance=10

\item We propose a self-calibrated training strategy to prevent the network from propagating the negative influence of error-prone pseudo labels. A mutual calibration loop is established between pseudo labels and network predictions to promote each other.

\item We open up a fresh perspective on that even a much smaller dataset (merely  $1.8$\% of ImageNet) with well-matched image-level annotations allows WSOD to achieve better performance. This encourages more existing data to be correctly annotated and further paves the way for the booming future of WSOD.

\item Our method outperforms existing WSOD methods on all metrics over five benchmark datasets, and meanwhile achieves averagely $94.7$\% performance of state-of-the-art fully supervised methods. We also demonstrate that our method retains its competitive edge on most metrics even without our proposed dataset.

\item We extend the proposed method to other fully supervised SOD methods. Our offered pseudo labels enable these methods to achieve comparatively high accuracy ($95.6$\% for BASNet \cite{Qin_2019_CVPR} and $97.3$\% for ITSD \cite{Zhou_2020_CVPR} on F-measure) while being free of pixel-level annotations, costing only $0.32$\% of labeling time for pixel-level annotation.
\end{itemize}

\section{Related Work}
\subsection{Salient Object Detection}

Early SOD methods mainly focus on detecting salient objects by utilizing handcraft features and setting various priors, such as center prior~\cite{jiang2013submodular}, boundary prior \cite{yang2013saliency} and so on \cite{zhu2014saliency, kim2014salient}. Recently, deep learning based methods demonstrate its advantages and achieve remarkable improvements. Plenty of promising works \cite{hou2017deeply, wu2019cascaded, SuLZXT19, ZhaoLFCYC19, WeiWH20} are proposed and present various effective architectures. Among them, Hou \etal \cite{hou2017deeply} present short connections to integrate the low-level and high-level features, and predict more detailed saliency maps. Wu \etal \cite{wu2019cascaded} propose a novel cascaded partial decoder framework and utilize generated relatively precise attention map to refine high-level features. 
in \cite{SuLZXT19, ZhaoLFCYC19}, researchers propose to explore boundary of the salient objects to predict a more detailed prediction. 
Although appealing performance these methods have achieved, vast high-quality pixel-level annotations are needed for training their models, which are time-consuming and laborious.

\begin{figure*}
\vspace{-0mm}
\includegraphics[width=1\linewidth]{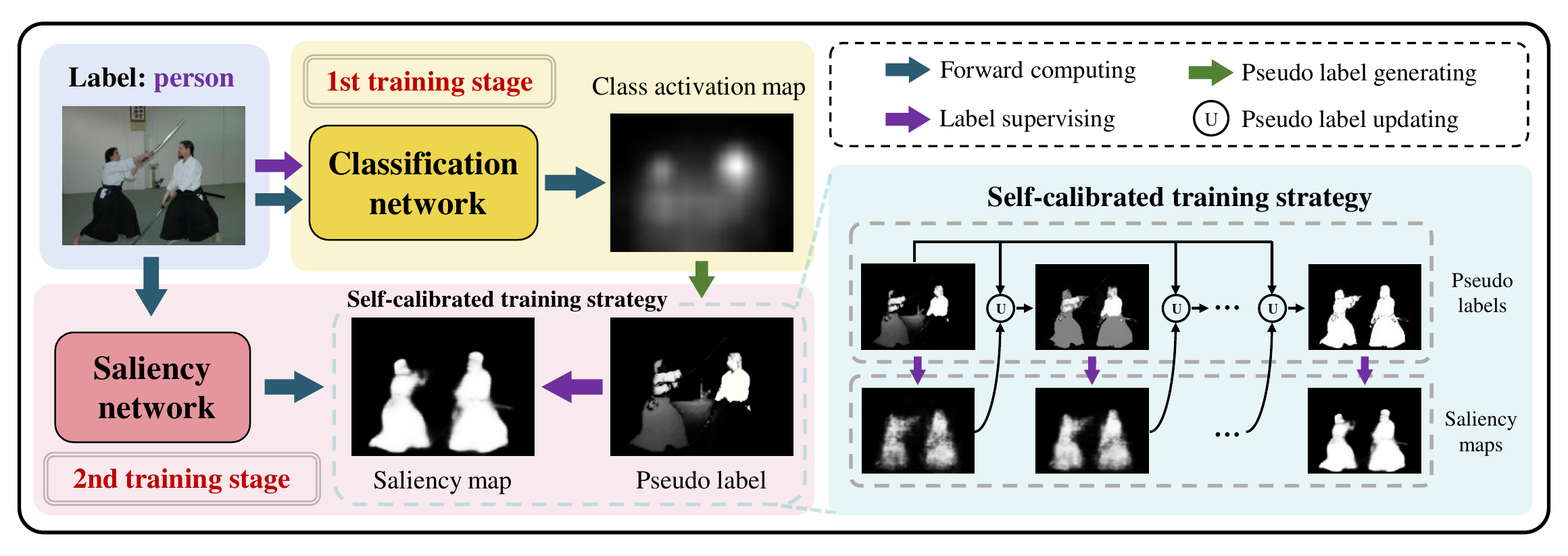}
\vspace{-5mm}
\caption{ Overall framework of our proposed method. In the first stage, classification labels are used to supervise classification network to generate CAMs and further produce pseudo labels. In the second stage, we train a saliency network with the above pseudo labels and propose a self-calibrated strategy to correct labels and predictions progressively.}
\label{overall}
\vspace{-2mm}
\end{figure*}

\subsection{Weakly Supervised Salient Object Detection}

For achieving a trade-off between labeling efficiency and model performance, researchers aim to perform salient object detect with low-cost annotations. 
To this end, WSOD is presented and achieves an appealing performance with image-level annotations only.

Wang \etal \cite{wang2017learning} design a foreground inference network (FIN) to predict saliency maps from image-level annotations, and introduce a global smooth pooling (GSP) to combine the advantages of global average pooling (GAP) and global max pooling (GMP), which explicitly computes the activation of salient objects. In \cite{li2018weakly}, Li \etal also perform WSOD based on image-level annotations, they adopt a recurrent self-training strategy and propose a conditional random field based graphical model to cleanse the noisy pixel-wise annotations by enhancing the spatial coherence as well as salient object localization. Based on a traditional method MB+ \cite{zhang2015minimum}, more accurate saliency maps are generated in less than one second per image. Zeng \etal \cite{zeng2019multi} intelligently utilize multiple annotations (\ie, classification and caption annotations) and design a multi-source weak supervision framework to integrate information from various annotations. Benefited from multiple annotations and an interactive training strategy, a really sample saliency network can also achieve appealing performance. 
All the above methods target to train a classification network (on existing large-scale multiple objects dataset, \ie, ImageNet \cite{imagenet_cvpr09} or Microsoft COCO \cite{lin2014microsoft}) to generate class activation maps (CAMs) \cite{zhou2016learning}, then perform different refinement methods to generate pseudo labels. Supervised by these pseudo labels directly, a saliency network is trained and predicts the final saliency maps.

Different from the aforementioned works, we argue that: 
\textbf{1)} Developing an effective training strategy encourages more accurate predictions even under the supervision of inaccurate pseudo labels which would mislead the networks.
\textbf{2)} Establishing accurate matches between classification labels and salient objects could facilitate the further development of WSOD.

\section{The Proposed Method}
In this section, we describe the details of our two-stage framework. As illustrated in the Figure {\color{red}\ref{overall}}, in the first training stage, we train a normal classification network based on the proposed saliency-based dataset, to generate more accurate pseudo labels. We then develop a saliency network using the pseudo labels in the second stage. A self-calibrated training strategy is proposed in this stage to immune network from inaccurate pseudo labels and encourage more accurate predictions.

\subsection{From Image-level to Pixel-level}

Class activation maps (CAMs) \cite{zhou2016learning} localize the most discriminative regions in an image using only a normal classification network and build a preliminary bridge from image-level annotations to pixel-level segmentation tasks.
In this paper, we adopt CAMs following the same setting of \cite{ahn2019weakly}, to generate pixel-level pseudo labels in the first training stage. To better understand our proposed approach, we will describe the generation of CAMs in a brief way. 

For a classification network, we discard all the fully connected layers and apply an extra global average pooling (GAP) layer as well 
as a convolution layer as previous works do. In the training phase, we take images in classification dataset as input, and compute its classification scores $Cls$ as follows:

\vspace{-0mm}
\begin{equation}
\begin{split}
Cls = {w_s}^T*GAP({F^5}) + {b_s},
\end{split}
\end{equation}
\vspace{-0mm}

\noindent where $F^5$ represents the features from the last convolution block, $GAP(\cdot)$ denotes the global average pooling operation and $w_s^T$ as well as $b_s$ are the learnable
 parameters of the convolution layer. In the inference phase, we compute the CAMs of images in DUTS-Train dataset as follows:

\vspace{-0mm}
\begin{equation}
\begin{split}
{C_{AM}} = \sum\limits_{k = 1}^N {{Cls_k}*Norm({\mathop{\rm Re}\nolimits} lu{({w_s}^T * {F^5} + {b_s}})_k}),
\end{split}
\end{equation}
\vspace{-0mm}

\noindent where $Relu(\cdot)$ and $Norm(\cdot)$ denote the relu activation function and normalization function, respectively. $w_s^T$ and $b_s$ are the shared parameters learnd in the training phase, $Cls_k$ represents the classification scores for category $k$ and $N$ represents the total number of categories. In this phase, multi-scale inference strategy is adopted, which rescales the original image into four sizes and computes the average CAMs as the final output. 

As Ahn \etal \cite{ahn2019weakly} have pointed out, CAMs mainly concentrate on the most discriminative regions and are too coarse to serve as pseudo labels. Various refinements have been conducted to generate pseudo labels. Different from \cite{wang2017learning, zeng2019multi} using an clustering algorithm SLIC \cite{achanta2012slic}, 
a plug-and-play module PAMR \cite{araslanov2020single} is adopt in our method. It performs refinement using the low-level color information of RGB images, which can be inserted into our framework flexibly and efficiently. Following the settings of \cite{wang2017learning, zeng2019multi}, we also adopt CRF \cite{krahenbuhl2011efficient} for a further refinement. Note that it is only used to generate pseudo labels in our method.

\subsection{Self-calibrated Training Strategy}
In the second training stage, a saliency network is trained with the pseudo labels generated in the first training stage. As is mentioned above, the relatively good results containing global representations of saliency is gradually degraded as the training process continues. 
A straightforward method to tackle this dilemma is setting a validation set to pick the best result during the training process. However, we argue that it may lead to sub-optimal results because 
\textbf{1)} despite good saliency representations are learned at the early training stage, the predictions are coarse and lack detail as the loss function is still converging (as shown in the $2^{nd}$ row in Figure {\color{red}\ref{introduction2}}). 
\textbf{2)} the capability of networks to learn saliency representation is not fully excavated. 
\textbf{3)} we believe that WSOD should not use any pixel-level ground truth in the training process, even as a validation set.
Following this main idea, we propose to establish a mutual calibration loop during the training process in which error-prone pseudo labels are recursively updated and calibrate network for better predictions in turn.

\begin{algorithm}[t]
\begin{algorithmic}[1]
\small
\vspace{2mm}
\caption{  Self-calibrated training strategy} 
\label{algorithm} 
\vspace{0.5mm}
\REQUIRE ~
The images from DUTS-Train dataset, $I_{n}$;~~
The predictions of saliency network, $P_n$;~~
The original pseudo labels generated in the $1_{st}$ stage, $Y_{1}$.
\ENSURE ~
the updated pseudo labels, $Y_{n+1}$.
\vspace{0.8mm}
\STATE Performing $2_{nd}$ training stage, maximum epoch is $N$.\\
\FOR{$n=1$ to $N$}
\STATE Refined predictions:  $P_n^{'}$ $=$ PAMR($P_n$, $I_n$);
\IF{ $P_n^{'}$$(x,y)$ $>$ $0.4$ } 
\vspace{0.5mm}
\STATE $P_n^{'}$$(x,y)$$ = 1$
\ELSE 
\STATE $P_n^{'}$$(x,y)$$ = 0$
\ENDIF 
\STATE weighting factor $\lambda$ = {$(n / N)^{0.5}$};\\
\STATE Updating pseudo labels: $Y_{n+1}$ $=$ $Y_{1}$ $ * (1-\lambda)$ $+$ $P_n^{'}$ $ *  \lambda$;
\ENDFOR

\end{algorithmic}
\end{algorithm}
\textbf{Insight:} As is discussed in the Section \uppercase\expandafter{\romannumeral1}, under the supervision of noisy pseudo labels, the saliency network goes from optimal to overfitting. On the one hand, in our weakly supervised settings, this “overfitting” manifestes itself as the network being affected by the noisy pseudo labels and learning the inaccurate noise information in them, which heavily restricts the performance of WSOD. It is also worth to mention that this is fundamentally different from the “overfitting” in supervised learning, the latter means that the network learns the biased information in a less comprehensive training set. On the other hand, we conclude reasons of the optimal point before overfitting as: 1) Although many pseudo labels are noisy and inaccurate, the whole pseudo labels still describe general saliency cues. It can provide a roughly correct guidance for the saliency network. 2) Before the loss converges, the saliency network is prone to learn the regular and generalized saliency cues rather than the irregular and noisy information in pseudo labels. Such kind of robustness is also discussed in \cite{FanZT20}. Motivated by the above analyses, we propose a self-calibrated training strategy to effectively utilize the robustness and tackle the negative overfitting.

To be specific, supervised by inaccurate pseudo labels $Y$, we take the predictions $P$ of the saliency network as saliency seeds. As is illustrated in Figure {\color{red}\ref{overall}}, coarse but more accurate seeds are predicted during the first few epochs regardless of the inaccurate supervision of error-prone pseudo labels. We take these seeds as correction terms to calibrate and update the original pseudo labels $Y$, while performing refinement again with PAMR. Detailed procedure is presentd in Algorithm {\color{red}\ref{algorithm}}, here we set a threshold to $0.4$ for the binarization operation on refined predictions $P^{'}$. We conduct self-calibrated strategy throughout the training process, that is, it is performed on each training batch. The loss function for this training stage can be described as:

\vspace{-1mm}
\begin{equation}
\begin{split}
L(P,Y) =  - \sum\limits_{i = 1}^{} {((1 - \lambda ) {y_i} + \lambda \, {p_i^{'}})*\log {p_i}} \\ - ((1 - \lambda )(1 - {y_i}) + \lambda (1 - {p_i^{'}}))*\log (1-{p_i}),
\end{split}
\end{equation}
\vspace{-0mm}

\noindent where $\lambda$ is the weighting factor that is illustrated in Algorithm {\color{red}\ref{algorithm}}. The intuition is that as the training process goes on, the saliency prediction is more accurate and larger weight should be given. $y_i$, $p_i$ and $p_i^{'}$ represent the elements of $Y$, $P$ and refined predictions $P^{'}$, respectively.

As is illustrated in the Figure {\color{red}\ref{overall}}, equipped with our proposed self-calibrated training strategy, inaccurate pseudo labels are progressively updated, and in turn supervise the network. This mutual calibration loop finally encourages both accurate pseudo labels and predictions.


\begin{figure}[t]
\vspace{-2mm}
\begin{center}
\includegraphics[width=1\linewidth]{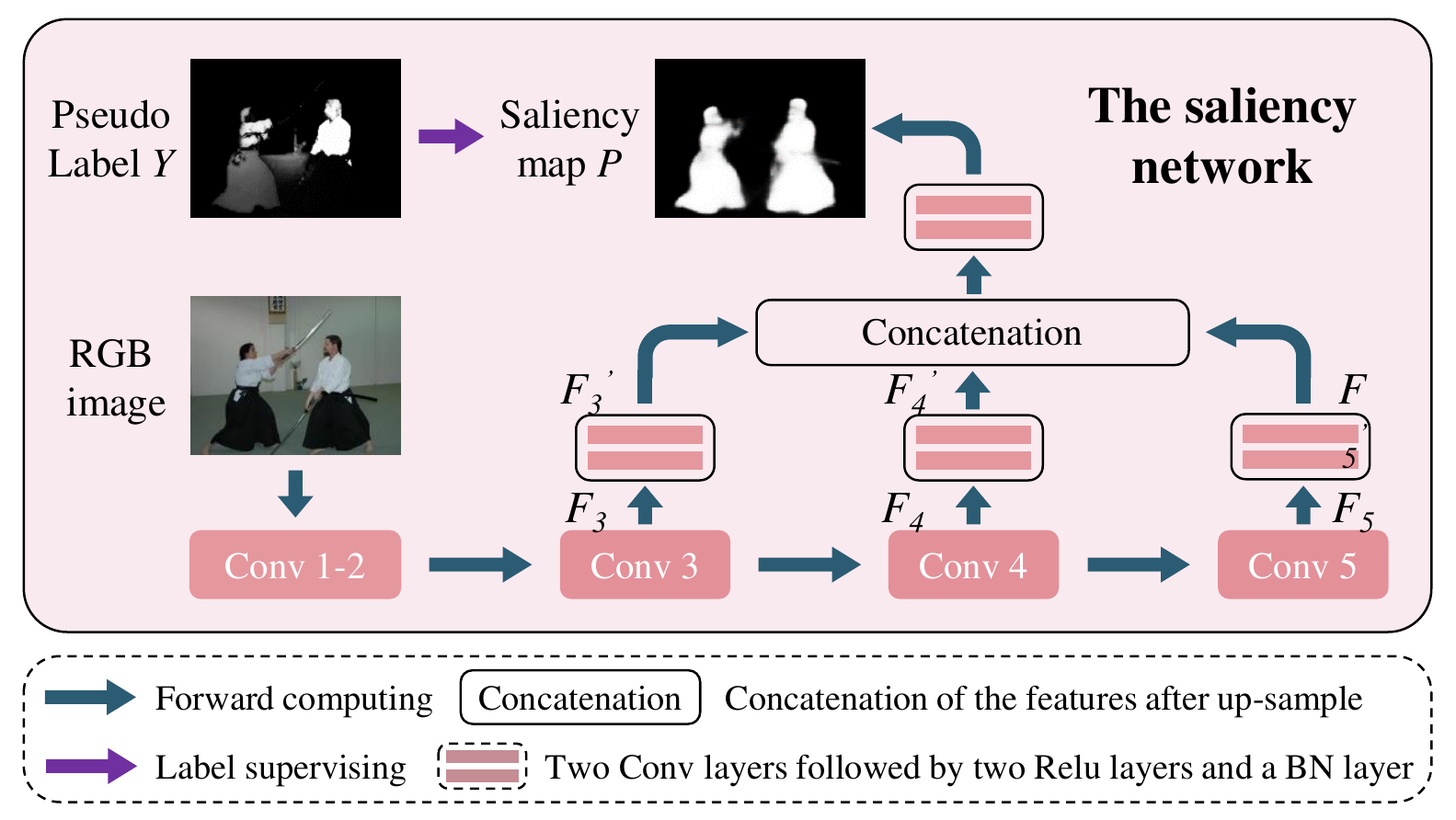}
\end{center}
\vspace{-4mm}
\caption{ Detailed structure of our saliency network. We adopt a simple encoder-decoder architecture and take prediction $P$ as our final result.}
\label{sal}
\vspace{-1mm}
\end{figure}

\subsection{Saliency Network}

As for the saliency network, we adopt a simple encoder-decoder architecture without any auxiliary modules, which is usually served as baseline for fully-supervised SOD methods \cite{hou2017deeply, wu2019cascaded}. As illustrated in Figure {\color{red}\ref{sal}}, for an image from DUTS-Train dataset, we take features $F_3$, $F_4$ and $F_5$ from the encoder, to generate $F_3^{'}$, $F_4^{'}$ and $F_5^{'}$ through two convolution layers, and then adopt a bottom-up strategy to perform feature fusion, which can be denoted as:

\vspace{-0mm}
\begin{equation}
{{P}}=\sigma (Conv(Cat(Up({{F}_{5}}^{'}),Up({{F}_{4}}^{'}),{{F}_{3}}^{'}))),
\end{equation}
\vspace{-1mm}

\noindent where $\sigma(\cdot)$ represents the sigmoid function, $Conv(\cdot)$ and $Cat(\cdot)$ denote the convolution and concatenation operation, respectively. $Up(\cdot)$ represents upsampling feature maps to the same size. 

In the decoder, the number of output channels of all the middle convolution layers are set to 64 for acceleration. 
Note that our final prediction $P$ is predicted in an end-to-end manner in the test phase without any post-processing.


\section{Dataset Construction}
To explore the advantages of accurate matches between image-level annotations and corresponding salient objects, 
we establish a saliency-based classification dataset, which ensures all the classification labels correspond to the salient objects. Following this main idea, we relabel an existing widely-adopted saliency training set DUTS-Train \cite{wang2017learning} with well-matched image-level annotations, namely DUTS-Cls dataset. It fits with WSOD better than existing large-scale classification datasets due to the accurate matches, and facilitates the further improvements for WSOD.

\begin{figure}
\vspace{0mm}
\begin{center}
\includegraphics[width=1\linewidth]{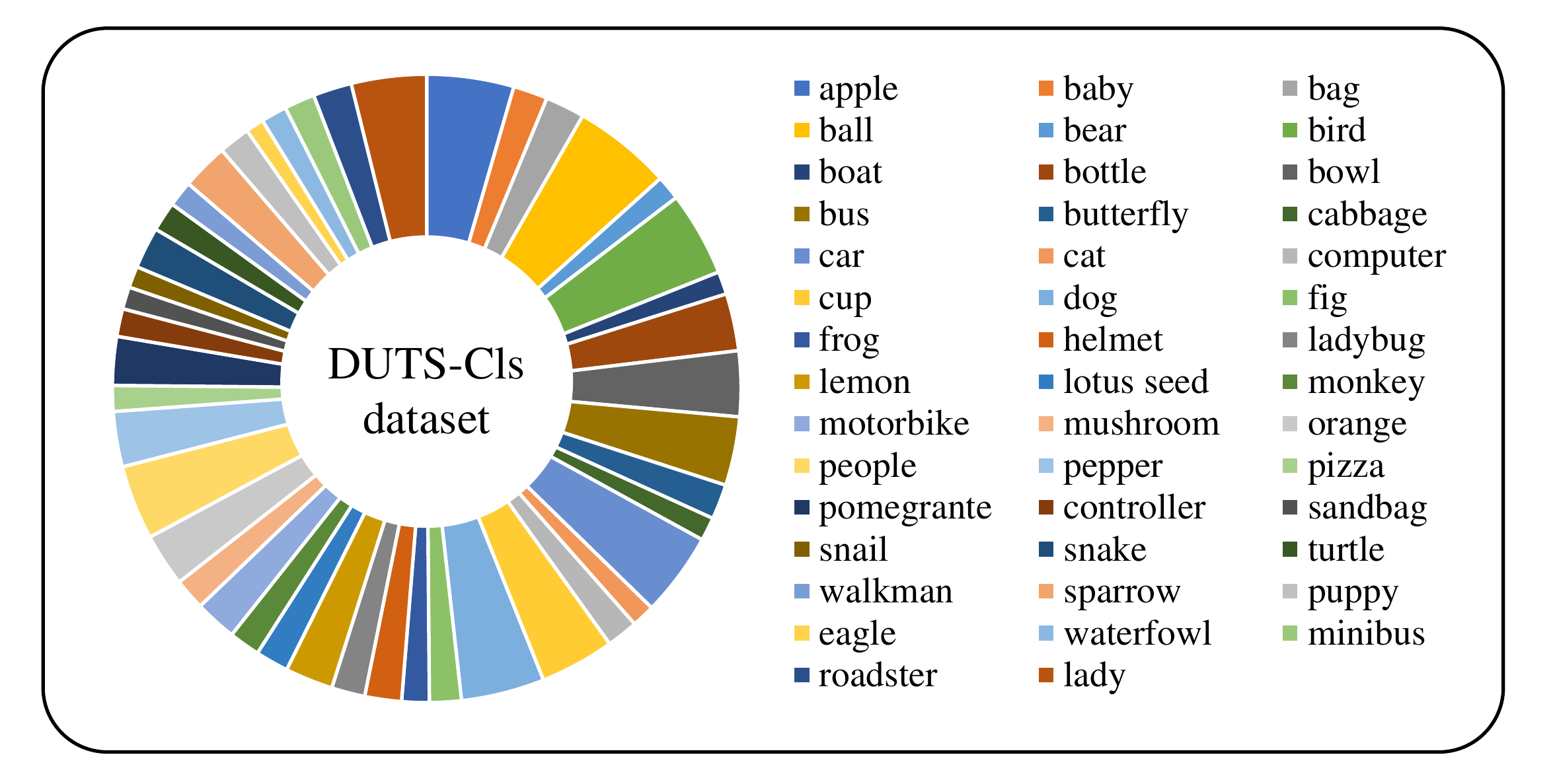}
\end{center}
\vspace{-4mm}
\caption{Our introduced DUTS-Cls is a saliency-based dataset with image-level annotations, containing 44 categories and 5959 images, in which all the classification labels correspond to the most salient objects in images.}
\label{dataset}
\vspace{-0mm}
\end{figure}

To be specific, we select and label images in DUTS-Train with image-level annotations, while discarding rare categories because only several images are contained. The proposed DUTS-Cls dataset contains 44 categories and 5959 images. As is illustrated in Figure {\color{red}\ref{dataset}}, it reaches a relative equilibrium in terms of image numbers of each category and covers most common categories.

\begin{table*}[!t]

	\centering
	\renewcommand\arraystretch{1.2}  
	\setlength{\abovecaptionskip}{0.6mm}
	\setlength{\belowcaptionskip}{0.6mm}
	\setlength{\tabcolsep}{0.4mm}
	\vspace{0mm}
	\begin{threeparttable}
	\caption{Quantitative comparisons of E-measure ($E_s$), S-measure ($S_{\alpha}$), F-measure ($F_{\beta}$) and MAE ($M$) metrics over five benchmark datasets. The supervision type (\textbf{Sup.}) I indicates using image-level annotations only, and I\&C represents developing WSOD on both image-level annotations and caption annotations simultaneously. \textbf{Num.} represents the number of training samples. - means unavailable results, Ours- and Ours represent our method trained on ImageNet and proposed DUTS-Cls dataset, respectively. The best two results are marked in \textbf{boldface} and {\textcolor{magenta}{magenta}}.}

	\label{quantitative}
    \begin{tabular}{ccp{0.7cm}<{\centering}p{0.7cm}<{\centering}p{0.7cm}<{\centering}p{0.7cm}<{\centering}p{0.7cm}<{\centering}p{0.7cm}<{\centering}p{0.7cm}<{\centering}p{0.7cm}<{\centering}p{0.7cm}<{\centering}p{0.7cm}<{\centering}p{0.7cm}<{\centering}p{0.7cm}<{\centering}p{0.7cm}<{\centering}p{0.7cm}<{\centering}p{0.7cm}<{\centering}p{0.7cm}<{\centering}p{0.7cm}<{\centering}p{0.7cm}<{\centering}p{0.7cm}<{\centering}p{0.7cm}<{\centering}}
    \toprule
    \multicolumn{1}{c}{\multirow{2.5}{*}{\textbf{Methods} }}&
    \multicolumn{1}{c}{\multirow{2.5}{*}{\textbf{Sup. }}}&
    \multicolumn{4}{c}{ECSSD}& \multicolumn{4}{c}{DUTS-Test}& \multicolumn{4}{c}{HKU-IS}& \multicolumn{4}{c}{DUT-OMRON}& \multicolumn{4}{c}{PASCAL-S}\cr
    \cmidrule(lr){3-6} \cmidrule(lr){7-10}\cmidrule(lr){11-14}\cmidrule(lr){15-18}\cmidrule(lr){19-22}
    &{}  &$S_{\alpha}$		&$E_s$		 &$F_{\beta}$ 	&$M$  	&$S_{\alpha}$		&$E_s$	 &$F_{\beta}$ 	&$M$  &$S_{\alpha}$	&$E_s$			 &$F_{\beta}$ 	&$M$ &$S_{\alpha}$ 	&$E_s$			 &$F_{\beta}$ 	&$M$  &$S_{\alpha}$		&$E_s$		 &$F_{\beta}$ 	&$M$\cr

	\midrule
	\multicolumn{1}{c}{\multirow{1}{*}{WSS~\cite{wang2017learning}}}
	&I 		
	&.811 	&.869 	&.823 	&.104 			&.748 	&.795 	&.654 	&.100  		&.822 	&.896 	&.821 	&.079 		&.725 	&.768 	&.603 	&.109 		&.744 	&.791 	&.715 	&.139		\cr  
	\multicolumn{1}{c}{\multirow{1}{*}{ASMO~\cite{li2018weakly}}}
	&I 			
	&.802 	&.853 	&.797 	&.110 			&.697 	&.772 	&.614 	&.116 			&- 		&- 		&- 		&- 			&.752 	&.776 	&.622 	&.101 		&.717 	&.772 	&.693 	&.149		\cr
	\multicolumn{1}{c}{\multirow{1}{*}{MSW~\cite{zeng2019multi}}}
	& I\&C 		
	&.827 		&.884 	&\textcolor{magenta}{.840} 	&.096		
	&.759 		&.814 	&.684 		&.091 		
	&.818 		&.895 	&.814 		&.084 		
	&\textcolor{magenta}{.756} 	&.763 	&.609 		&.109 		
	&.768 		&.790 	&.713 		&.133		\cr

	\midrule   

	\multicolumn{1}{c}{\multirow{1}{*}{Ours-}}	
	&I	 
	&\textcolor{magenta}{.836} 	&\textcolor{magenta}{.887} 	&.838 								&\textcolor{magenta}{.083}		
	&\textcolor{magenta}{.770} 	&\bf{.830} 						&\bf{.689} 						&\textcolor{magenta}{.079} 	
	&\textcolor{magenta}{.836} 	&\textcolor{magenta}{.907} 	&\textcolor{magenta}{.822} 	&\textcolor{magenta}{.064} 	
	&.743 								&\textcolor{magenta}{.807} 	&\textcolor{magenta}{.643} 	&\textcolor{magenta}{.085} 	
	&\textcolor{magenta}{.778} 	&\textcolor{magenta}{.818} 	&\textcolor{magenta}{.742} 	&\textcolor{magenta}{.111}		\cr
 
	\multicolumn{1}{c}{\multirow{1}{*}{Ours}}	
	&I	
	&\bf{.858} 	&\bf{.901} 	&\bf{.853} 	&\bf{.071}		
	&\bf{.776} 	&\textcolor{magenta}{.829} 						&\textcolor{magenta}{.688}						&\bf{.077} 	
	&\bf{.850} 	&\bf{.918} 	&\bf{.835} 	&\bf{.058} 	
	&\bf{.766} 	&\bf{.817} 	&\bf{.667} 	&\bf{.078} 	
	&\bf{.781} 	&\bf{.824} 	&\bf{.749} 	&\bf{.108}		\cr

	\bottomrule
    \end{tabular}
    \end{threeparttable}
    \vspace{-2mm}
\end{table*}

\begin{figure*}[!t]
\vspace{2mm}
\includegraphics[width=1.00\linewidth]{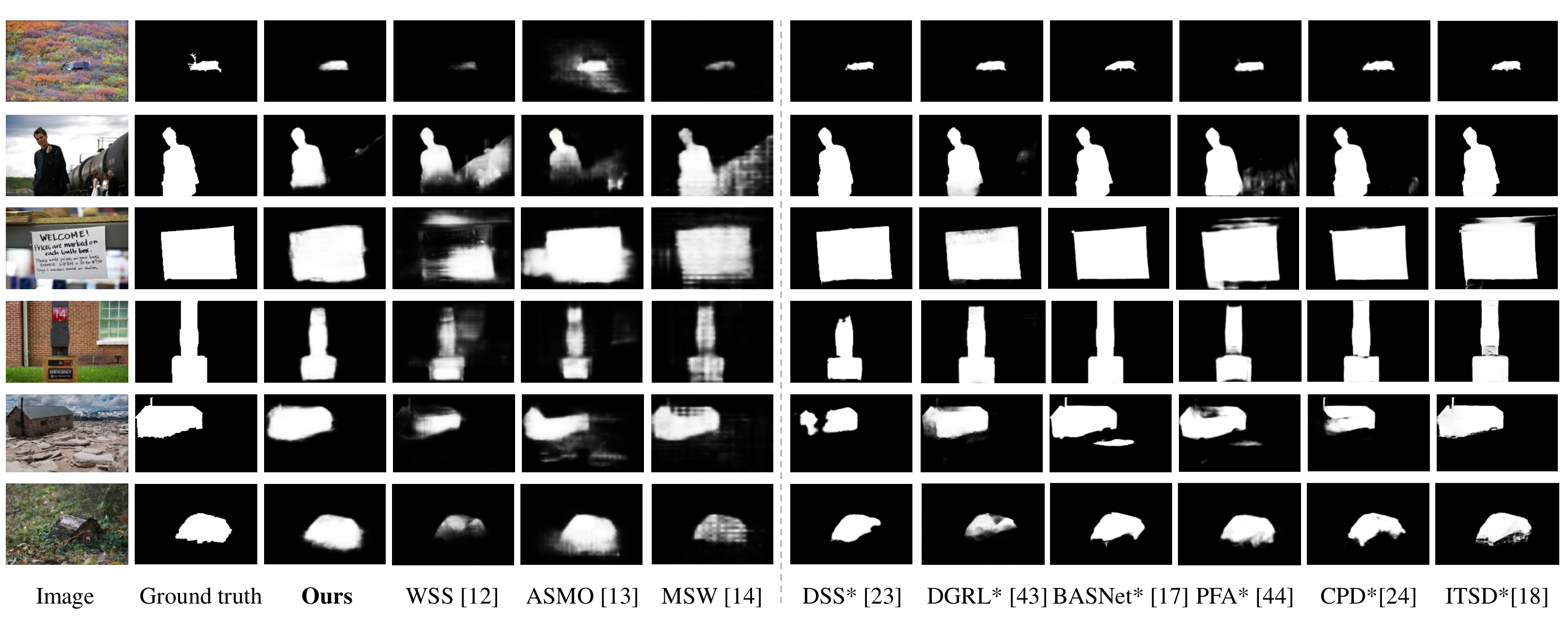}
\setlength{\abovecaptionskip}{0.2cm}
\setlength{\belowcaptionskip}{0.2cm}
\vspace{-5mm}
\caption{Visual comparisons of our method with existing WSOD methods as well as six state-of-the-art fully supervised SOD methods (marked with *) in some challengling scenes.}
\label{qualitative}
\vspace{-2mm}
\end{figure*}

It is worth mentioning that labeling image-level annotations is quite fast, which only takes less than 1 seconds per image. Compared to about 3 minutes \cite{niu2012leveraging} for labeling a pixel-level ground truth, it takes less than $0.56$\% of the time and labor cost for a sample. Annotating DUTS-Cls dataset (5959 samples) only costs $0.32$\% of labeling time than annotating the whole DUTS-Train dataset (10553 samples) with pixel-level ground truth. This indicates that exploring WSOD with image-level annotation is quite efficient.
Moreover, the DUTS-Cls dataset with well-matched image-level annotations offers a better choice for WSOD than ImageNet, and we genuinely hope it could contribute to the community
and encourage more existing data to be correctly annotated at image level.

\section{Experiments}

\subsection{Implementation Details}
We implement our method on the Pytorch toolbox with a single RTX 2080Ti GPU. The backbone adopted in our method is DenseNet-169~\cite{Huang2017DenselyCC}, which is same as the latest work~\cite{zeng2019multi}. During the first training stage, we train a classification network on our proposed DUTS-Cls dataset. In this stage, we adopt the Adam optimization algorithm \cite{kingma2014adam}, the learning rate is set to 1e-4 and maximum epoch is set to $20$. In the second training stage, we only take the RGB images from DUTS-Train as our training set. In this stage, we use Adam optimization algorithm with the learning rate 3e-6 and maximum epoch 25. The batch size of both training stages is set to $20$ and all the training and testing images are resized to $256\times 256$. 


\begin{figure}
\vspace{0mm}
\includegraphics[width=1.00\linewidth]{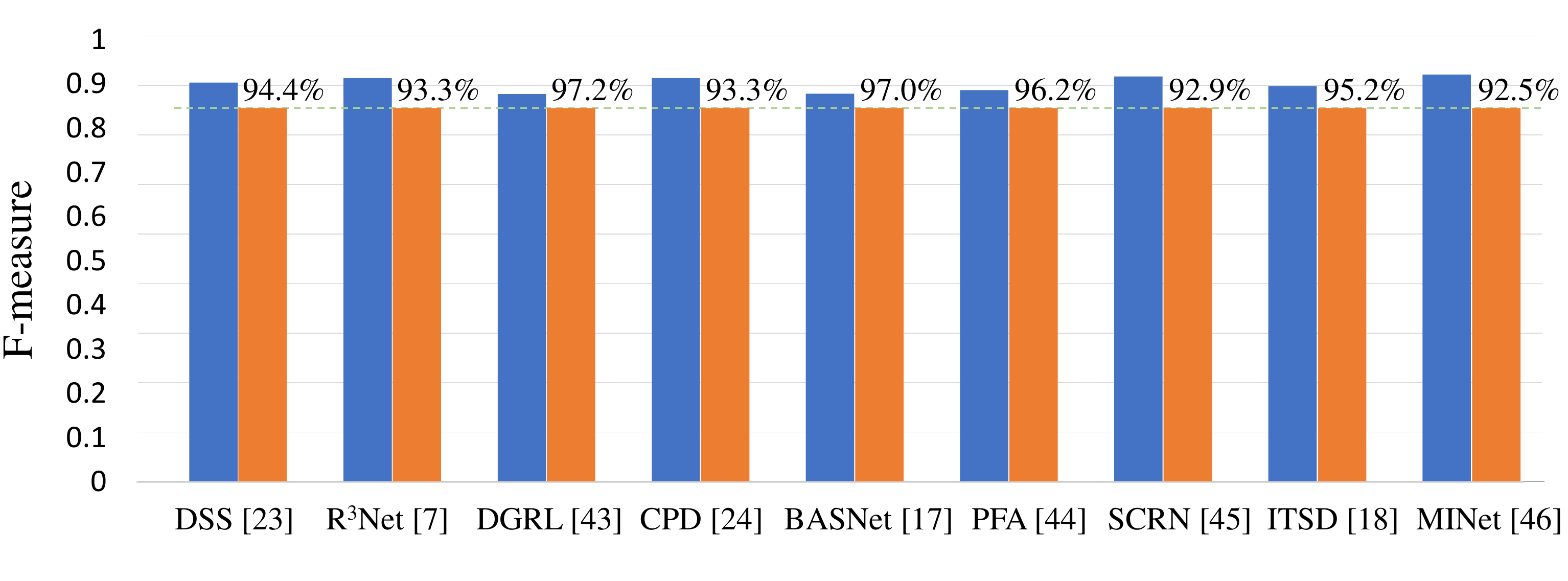}
\setlength{\abovecaptionskip}{0.2cm}
\setlength{\belowcaptionskip}{0.2cm}
\vspace{-5mm}
\caption{ Comparison of our method with 9 fully supervised methods on ECSSD dataset. The blue column represents the performance of each fully supervised methods and the orange one indicates ours.
The corresponding data denote the percentages of performance of our method in different fully supervised methods.}
\label{with fully}
\vspace{-2mm}
\end{figure}

\noindent \textbf{Hyperparameters setting.} For the weighting factor $\lambda$ of self-calibrated strategy, we conduct hyper-parameter experiments on ECSSD \cite{yan2013hierarchical} dataset to pick the optimal value through F-measure \cite{achanta2009frequency}. According to the results (0.5 to 0.848, 0.6 to 0.853 and 0.7 to 0.849), we finally set the hyper-parameter $\lambda$ to 0.6.

\subsection{Datasets and Evaluation Metrics}
For a fair comparison, we train our model on ImageNet and our proposed DUTS-Cls dataset respectively, the results are shown in Table {\color{red}\ref{quantitative}}.
We conduct comparisons on five following widely-adopted test datasets. ECSSD \cite{yan2013hierarchical}: contains 1000 images which cover various scenes. DUT-OMRON \cite{yang2013saliency}: includes 5168 challenging images consisting of single or multiple salient objects with complex contours and backgrounds. PASCAL-S \cite{Li_2014_CVPR}: is collected from the validation set of the PASCAL VOC semantic segmentation dataset \cite{everingham2010pascal}, and contains 850 challenging images. HKU-IS \cite{li2015visual}: includes 4447 images, many of which contain multiple disconnected salient objects. DUTS \cite{wang2017learning}: is the largest salient object detection benchmark, which contains 10553 training samples (DUTS-Train) and 5019 testing samples (DUTS-Test). Most images in DUTS-Test are challenging with various locations and scales.

To evaluate our method in a comprehensive and reliable way, we adopt four well-accepted metrics, including S-measure \cite{fan2017structure}, E-measure \cite{fan2018enhanced}, F-measure \cite{achanta2009frequency} as well as Mean Absolute Error (MAE).

\subsection{Comparison with State-of-the-arts}
We compare our method with all the existing image-level annotation based WSOD methods: WSS \cite{wang2017learning}, ASMO \cite{li2018weakly} and MSW \cite{zeng2019multi}.
To further demonstrate the effectiveness of our weakly supervised methods, we also compare the proposed method with nine state-of-the-art fully supervised methods including DSS \cite{hou2017deeply}, R$^{3}$Net \cite{deng2018r3net}, DGRL \cite{wang2018detect}, BASNet \cite{Qin_2019_CVPR}, PFA \cite{zhao2019pyramid}, CPD \cite{wu2019cascaded}, SCRN \cite{Wu_2019_ICCV}, ITSD \cite{Zhou_2020_CVPR} and MINet \cite{MINet-CVPR2020}, all of which are trained on pixel-level ground truth and based on DNNs. For a fair comparison, we use the saliency maps provided by authors and perform the same evaluation code for all methods.

\noindent \textbf{Quantitative evaluation.} Table {\color{red}\ref{quantitative}} shows the quantitative comparison on four evaluation metrics over five datasets. It can be seen that our method outperforms all the weakly supervised methods on all metrics. Especially, $31.0$\% improvement on HKU-IS and $28.4$\% on DUT-OMRON are achieved on MAE metric.
Our method also improves the performance on two challenging datasets DUT-ORMON and PASCAL-S by a large margin, which indicates that our method can explore more accurate saliency cues even in complex scenes. 
\textbf{Additionally}, the proposed saliency-based dataset with well-matched image-level annotations enables our method to achieve better performance, while far less training samples (less than $1.45$\% of the latest work MSW~\cite{zeng2019multi}) are required. 
To prove the effect of our method in a more objective manner, we also train our method on ImageNet dataset following the previous works. The results of "{Ours-}" shown in Table {\color{red}\ref{quantitative}} demonstrate that our method can outperform existing methods on most metrics even without the proposed dataset thanks to the effective strategy.
\textbf{Moreover}, we also compare our method with nine state-of-the-art fully supervised methods.
It can be seen in Figure {\color{red}\ref{with fully}} that our method, even with the image-level annotations only and a simple baseline network without any auxiliary modules, can also achieve $94.7$\% accuracy of fully supervised methods on average.

\noindent \textbf{Qualitative evaluation.} In Figure {\color{red}\ref{qualitative}}, we show the qualitative comparisons of our method with existing three WSOD methods as well as six state-of-the-art fully supervised methods. 
It can be seen that our method could discriminate salient objects from various challenging scenes (such as small objects case in the $1^{st}$ row and complex background cases in the $2^{nd}$ and $3^{rd}$ rows) and achieve more complete and accurate predictions.
\textbf{Moreover}, compared with the fully supervised methods, our method also predicts comparable and even better results in some cases, such as the complete house and log in the $5^{th}$ and $6^{th}$ rows. 
But we would like to point out that our results also need to be improved in term of the boundary of the salient objects. 

\begin{table*}[!t]
	\renewcommand\arraystretch{1.2}  
  	\centering
  	\setlength{\tabcolsep}{0.78mm}
  	\setlength{\abovecaptionskip}{0.0cm}
	\setlength{\belowcaptionskip}{0.6cm}
	\vspace{0mm}
  	\begin{threeparttable}
	\caption{ Quantitative results of ablation studies. \textbf{Dataset} represents different training sets used in the first training stage. \textbf{Strategy} denotes training strategy used in the second stage, - indicates the baseline model without any training strategy and +SC represents adopting our proposed self-calibrated strategy.}
	\label{ablation}
	\begin{tabular}{ccp{1.5cm}<{\centering}p{1.2cm}<{\centering}p{1.2cm}<{\centering}p{1.0cm}<{\centering}p{1.0cm}<{\centering}p{1.0cm}<{\centering}p{1.0cm}<{\centering}p{1.0cm}<{\centering}p{1.0cm}<{\centering}p{1.0cm}<{\centering}p{1.0cm}<{\centering}p{1.0cm}<{\centering}p{1.0cm}<{\centering}p{1.0cm}<{\centering}}

	\toprule
    \multicolumn{0}{c}{  }&
    \multicolumn{2}{c}{\multirow{1}{*}{\bf Dataset}}&
    \multicolumn{2}{c}{\multirow{1}{*}{\bf Strategy}}&
    \multicolumn{2}{c}{ECSSD}& 
    \multicolumn{2}{c}{DUTS-Test}& 
    \multicolumn{2}{c}{HKU-IS} & 
	 \multicolumn{2}{c}{DUT-OMRON}& 
	 \multicolumn{2}{c}{PASCAL-S}\cr
    \cmidrule(lr){2-3} \cmidrule(lr){4-5} \cmidrule(lr){6-7} \cmidrule(lr){8-9} \cmidrule(lr){10-11} \cmidrule(lr){12-13} \cmidrule(lr){14-15} 
    &{\small ImageNet} &{\small DUTS-Cls} &{-} &{+ SC} &$F_{\beta}$$\uparrow$ 	&$M$$\downarrow$ 		&$F_{\beta}$$\uparrow$ 	&$M$$\downarrow$   		&$F_{\beta}$$\uparrow$ 	&$M$$\downarrow$     		&$F_{\beta}$$\uparrow$ 	&$M$$\downarrow$     		&$F_{\beta}$$\uparrow$ 	&$M$$\downarrow$  \cr

	\midrule

	& \checkmark & 					& \checkmark	& 				& 0.776		& 0.121		& 0.642		& 0.094		& 0.773		& 0.090		& 0.568		& 0.111		& 0.694		& 0.140	\cr
   
	& 				& \checkmark		& \checkmark	& 				& 0.836 	& 0.096 	& 0.675		& 0.085		& 0.822		& 0.075		& 0.648  	& 0.083		& 0.735		& 0.126	\cr

	& \checkmark	& 					& 				& \checkmark	& 0.838		& 0.083		& \bf0.689	& 0.079		& 0.822		& 0.064		& 0.643		& 0.085		& 0.742		& 0.111	\cr
  
	& 				& \checkmark		& 				& \checkmark	&\bf0.853	&\bf0.071	& 0.688		&\bf0.077	&\bf0.835	&\bf0.058	&\bf0.667	&\bf0.078	&\bf0.749	&\bf0.108	\cr
	\bottomrule
	\end{tabular}
	\end{threeparttable}
	\vspace{-2mm}
\end{table*}

\subsection{Ablation Studies}

\begin{figure}
\vspace{2mm}
\includegraphics[width=1.00\linewidth]{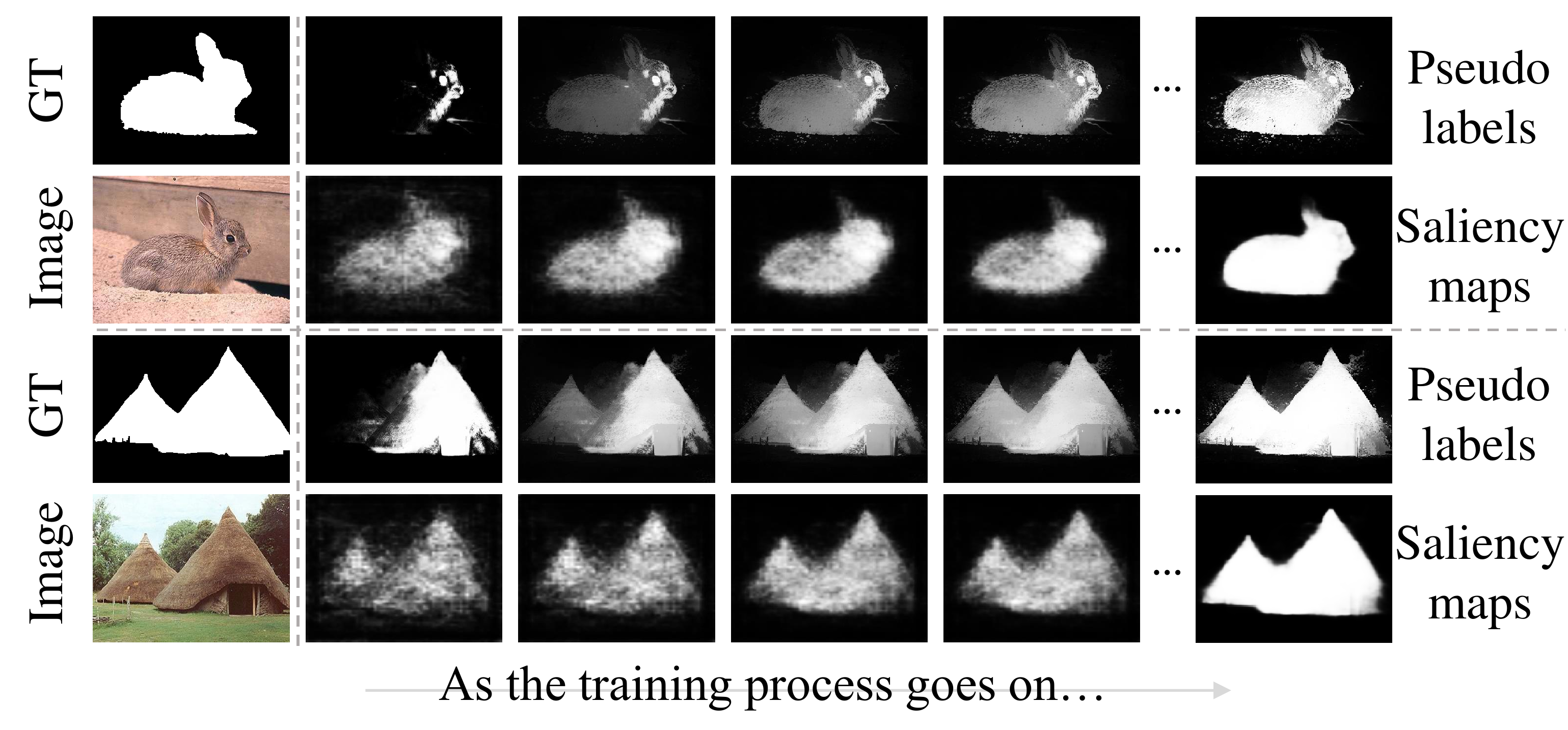}
\setlength{\abovecaptionskip}{0.2cm}
\setlength{\belowcaptionskip}{0.2cm}
\vspace{-5mm}
\caption{ Visual analysis of the effectiveness of our proposed self-calibrated strategy during the training process, noting that the ground truth is just for exhibition and not used in our framework.}
\label{loss}
\vspace{2mm}
\end{figure}

\noindent \textbf{Effect of the self-calibrated strategy.} We conduct experiments on both ImageNet ($1^{st}$ and $3^{rd}$ rows) and DUTS-Cls ($2^{nd}$ and $4^{th}$ rows) settings in Table {\color{red}\ref{ablation}}. It can be seen that the proposed self-calibrated strategy can not only enhance the performance of our method on ImageNet setting greatly, but also achieve great improvements even on the DUTS-Cls setting, especially on MAE metrics.
\textbf{Besides}, the effectiveness of the proposed self-calibrated strategy can also be demonstrated by the visual results in Figure {\color{red}\ref{loss}}. It can be seen that the proposed strategy can keep and enhance the globally good representations during the training process, and predict accurate saliency maps even supervised by error-prone pseudo labels.
\textbf{Moreover}, for a comprehensive evaluation, 1) We change the pseudo label by using two traditional SOD methods BSCA~\cite{qin2015saliency} and MR \cite{yang2013saliency}, and then train our model with and without the proposed strategy respectively, the results are shown in the first four rows in Table {\color{red}\ref{scloss}}. 2) We further apply our method on the lasted work MSW \cite{zeng2019multi} by just adding our proposed strategy in the last two rows in Table {\color{red}\ref{scloss}}. These results strongly prove that the self-calibrated strategy can not only work well on our method, but also effective for other pseudo labels and other works.


\begin{table}[!t]
	\renewcommand\arraystretch{1.1}  
  	\centering
  	\setlength{\tabcolsep}{0.8mm}
  	\setlength{\abovecaptionskip}{0.0cm}
	\setlength{\belowcaptionskip}{0.6cm}
	\vspace{0mm}
  	\begin{threeparttable}
	\caption{The effectiveness of our proposed self-calibrated strategy on ECSSD dataset. \textbf{+ SC} indicates simply applying our self-calibrated strategy during the training process.}
	\label{scloss}
	\begin{tabular}{cccp{1.3cm}<{\centering}p{1.1cm}<{\centering}p{1.1cm}<{\centering}p{1.1cm}<{\centering}p{1.1cm}<{\centering}}

	\toprule
	\multicolumn{2}{c}{ \textbf{ Method } $  $}&
	\multicolumn{1}{c}{ \textbf{Strategy}}&
	\multicolumn{1}{c}{ $S_{\alpha}$$\uparrow$}&
	\multicolumn{1}{c}{ $E_s$$\uparrow$}&
	\multicolumn{1}{c}{ $F_{\beta}$$\uparrow$}&
	\multicolumn{1}{c}{ MAE $\downarrow$}\cr

	\midrule
	\multicolumn{2}{c}{\multirow{2}{*}{BSCA~\cite{qin2015saliency}}}    
	& - 				& 0.846 		& 0.884 		& 0.814 		& 0.084		\cr
	& & + SC 			& \bf+0.007 	& \bf+0.009 	& \bf+0.018 	& \bf-0.008	\cr
	\midrule
	\multicolumn{2}{c}{\multirow{2}{*}{MR~\cite{yang2013saliency}}}   
	& -   				& 0.839 		& 0.884 		& 0.823 		& 0.085		\cr
	& & + SC 			& \bf+0.014	& \bf+0.010 	& \bf+0.016 	& \bf-0.009	\cr
	\midrule
	\multicolumn{2}{c}{\multirow{2}{*}{MSW \cite{zeng2019multi}}}   
	& -  				& 0.827 		& 0.884 		& 0.840 		& 0.096		\cr
	& & + SC 			& \bf+0.017	& \bf+0.012 	& \bf+0.014 	& \bf-0.014	\cr
	\bottomrule

	\end{tabular}
	\end{threeparttable}
	
\end{table}

\begin{figure}[t]
\vspace{2mm}
\includegraphics[width=1\linewidth]{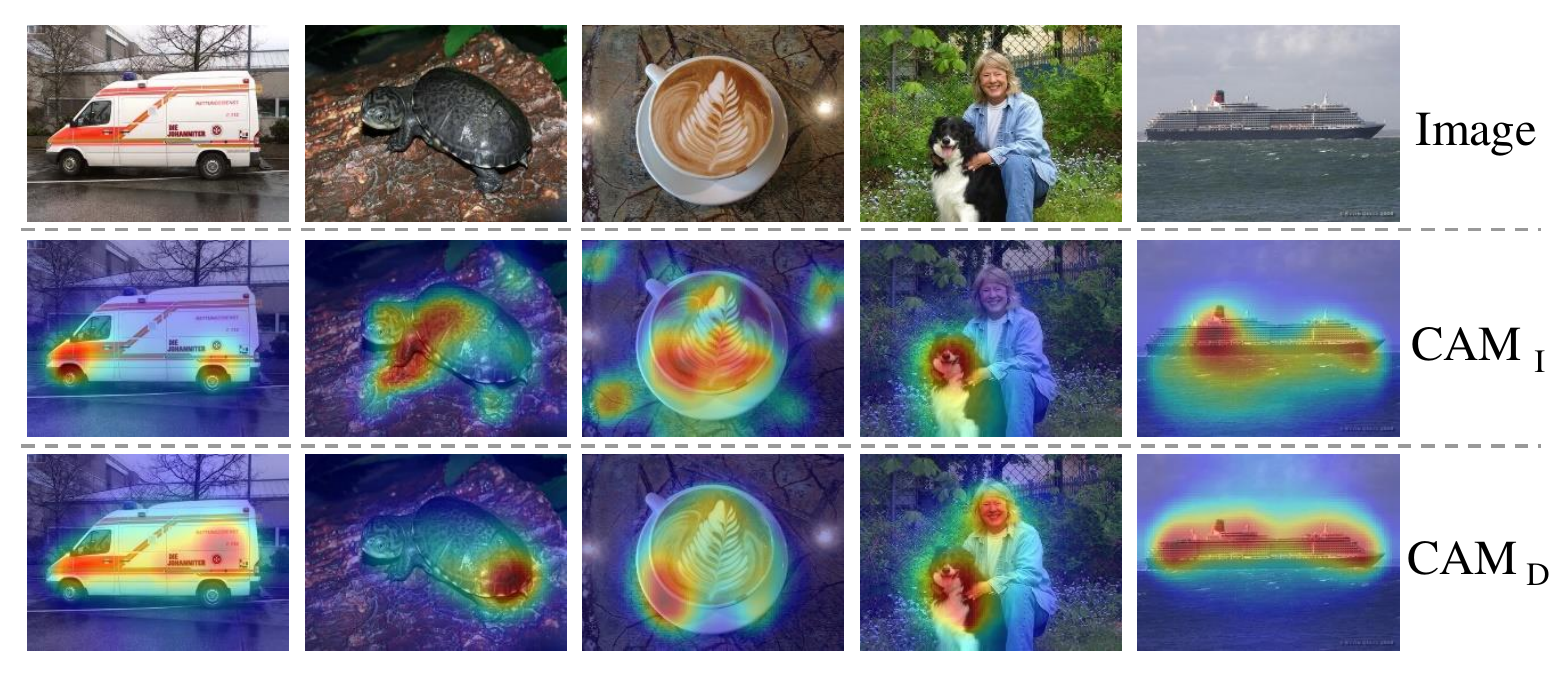}
\setlength{\abovecaptionskip}{0.2cm}
\setlength{\belowcaptionskip}{0.2cm}
\vspace{-5mm}
\caption{Visual analysis of effect of DUTS-Cls datset. CAM$_I$ and CAM$_D$ represent the CAMs generated by training on ImageNet and our DUTS-Cls dataset, respectively. Heatmap is adopted for better visualization.}
\label{cam}
\vspace{-2mm}
\end{figure}

\begin{figure}
\vspace{2mm}
\begin{center}
\includegraphics[width=1\linewidth]{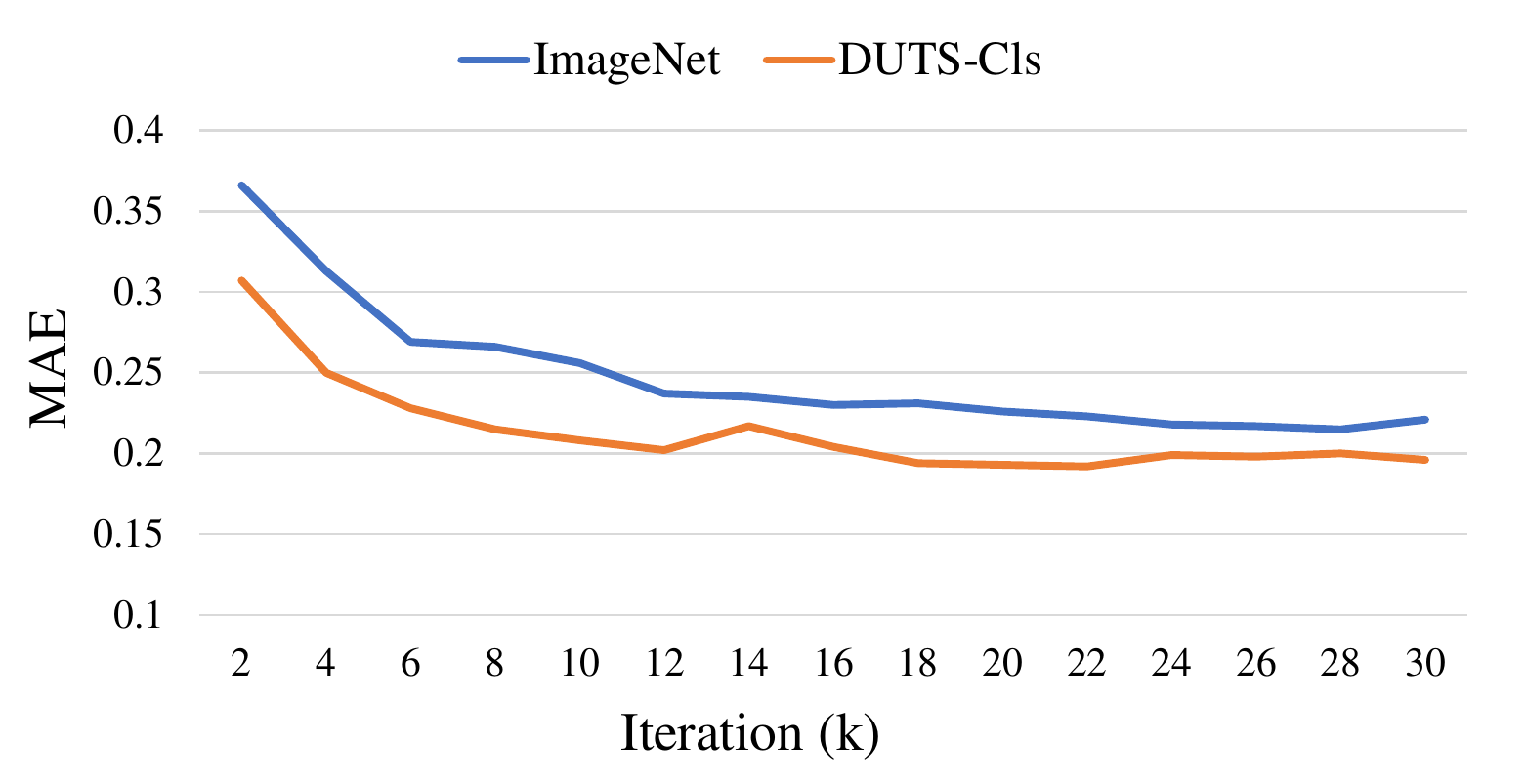}
\end{center}
\vspace{-4mm}
\caption{ Experiments on the effect of our proposed DUTS-Cls dataset. We conduct experiments on the classification branch of the latest work MSW \cite{zeng2019multi} for a fair comparison, the results are tested on the ECSSD dataset.}
\vspace{-0mm}
\label{sota}
\end{figure}

\noindent \textbf{Effect of the DUTS-Cls dataset.} We introduce a saliency-based dataset with well-matched image-level annotations to offer a better choice for WSOD.
The first two rows in Table {\color{red}\ref{ablation}} demonstrate that DUTS-Cls dataset encourages the baseline model to achieve remarkable improvements, compared to ImageNet dataset. 
And as is illustrated in the last two rows in Table {\color{red}\ref{ablation}}, it also proves its superiority by a steady improvement on most metrics even if good enough performance is already achieved by adopting the self-calibrated strategy. This is consistent with our argument that the cross-domain inconsistency does impede the performance of WSOD, and a saliency-based dataset can settle this matter better. 
\textbf{Additionally}, we visualize the CAMs trained on ImageNet (named CAM$_I$) and DUTS-Cls (named CAM$_D$) in Figure {\color{red}\ref{cam}}, it can be seen that CAM$_D$ have higher activation level within the salient objects trained on well-matched DUTS-Cls dataset.
\textbf{Last but not least}, to further prove the effectiveness of the proposed DUTS-Cls dataset objectively, we also train the latest work MSW \cite{zeng2019multi} on the DUTS-Cls dataset.
As is shown in Figure {\color{red}\ref{sota}}, by simply replacing ImageNet with DUTS-Cls, considerable improvements are achieved in less training iterations.
It is worth to mention that the DUTS-Cls dataset reaches less than $1.8$\% percent of ImageNet in terms of sample size. This strongly demonstrates the effectiveness and generalizability of the well-matched DUTS-Cls dataset for WSOD.

\subsection{Effectiveness on Unseen Category}

The category number of classification dataset inevitably influences the performance of WSOD. Unlike ImageNet including 200 various categories, our proposed DUTS-Cls dataset only contains 44 categories. It is necessary to evaluate the effectiveness of our method as well as DUTS-Cls dataset on unseen categories.

To this end, we choose THUR \cite{cheng2014salientshape} as the benchmark dataset for this experiment. THUR is a high-quality saliency dataset which consists of five categories including butterfly, coffee mug, dog, giraffe and airplane. The category airplane is unseen to our DUTS-Cls dataset but seen to ImageNet, while the category giraffe is unseen to both ImageNet and DUTS-Cls dataset. As is illustrated in the $2^{nd}$ and $3^{rd}$ rows in Table {\color{red}\ref{unseen}}, 
DUTS-Cls dataset encourages better predictions on the whole THUR dataset, and also outperforms ImageNet by a large margin on both airplane and giraffe categories.
It proves the generalizability and effectiveness of the proposed DUTS-Cls dataset. 
Besides, the superiority of our method on unseen categories can be demonstrated in the $1^{st}$ and $2^{nd}$ rows of Table {\color{red}\ref{unseen}}.
Moreover, except the airplane and giraffe categories, our method also behaves well on other various unseen categories such as the cases shown in Figure {\color{red}\ref{qualitative}}. It further supports the effect of our method on unseen categories.

\begin{table}[!t]
	\renewcommand\arraystretch{1.3}  
	\small
  	\centering
  	\setlength{\tabcolsep}{0.25mm}
  	\setlength{\abovecaptionskip}{0.0cm}
	\setlength{\belowcaptionskip}{0.6cm}
	\vspace{0mm}
  	\begin{threeparttable}
	\caption{The quantitative results of effectiveness on unseen category. \textbf{Dataset} represents the training set used in the first training stage, THUR-plane and THUR-giraffe denote the samples of plane and giraffe in THUR dataset, respectively.}
	\label{unseen}
	\begin{tabular}{p{1.3cm}<{\centering}p{1.1cm}<{\centering}p{0.9cm}<{\centering}p{0.9cm}<{\centering}p{0.9cm}<{\centering}p{0.9cm}<{\centering}p{0.9cm}<{\centering}p{0.9cm}<{\centering}}

	\toprule
    \multicolumn{1}{c}{\multirow{2.5}{*}{\bf Method }}&
    \multicolumn{1}{c}{\multirow{2.5}{*}{\bf  Dataset }}&
    \multicolumn{2}{c}{THUR}& \multicolumn{2}{c}{THUR-plane}& \multicolumn{2}{c}{THUR-giraffe} \cr
    \cmidrule(lr){3-4} \cmidrule(lr){5-6} \cmidrule(lr){7-8}
    &{}  &$F_{\beta}$$\uparrow$ 	&$M$$\downarrow$ 		&$F_{\beta}$$\uparrow$ 	&$M$$\downarrow$   		&$F_{\beta}$$\uparrow$ 	&$M$$\downarrow$  \cr

	\midrule
	\multicolumn{2}{c}{\multirow{1}{*}{MSW \cite{zeng2019multi}}}  {\multirow{1}{*}{ ImageNet}} 
	& 0.624 			& 0.104 		& 0.716  		&  0.079	 	& 0.547 		& 0.088 		\cr
	\midrule
   \multicolumn{2}{c}{\multirow{2}{*}{Ours}}  {\multirow{1}{*}{ ImageNet}}   
	& 0.676 			& 0.089 		& 0.788 		& 0.055 		& 0.550 		& 0.088		\cr
	\multicolumn{2}{c}{\multirow{2}{*}{$ $}} {\multirow{1}{*}{ {$ $DUTS-Cls}}} 
	&\bf0.689 		& \bf0.082 	& \bf0.809 	& \bf0.050 	& \bf0.588 	& \bf0.073 	\cr

	\bottomrule
	\end{tabular}
	\end{threeparttable}
	\vspace{-0mm}
\end{table}


\subsection{Applications}

\begin{figure}
\vspace{-0mm}
\begin{center}
\includegraphics[width=1\linewidth]{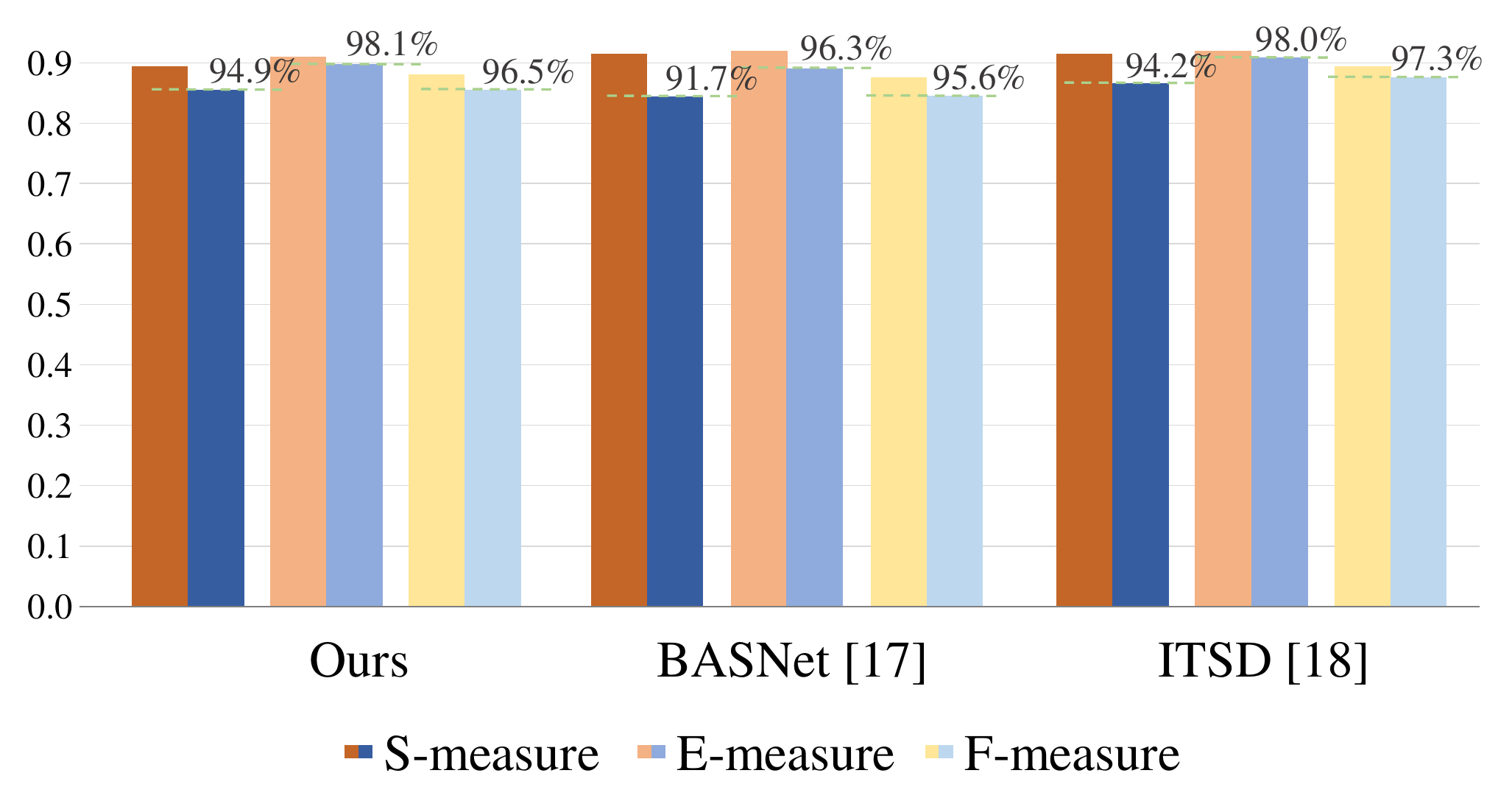}
\end{center}
\vspace{-4mm}
\caption{ Comparisons of different methods trained on our offered labels (the right one) and ground truth (the left one) on ECSSD dataset. The number on each data pair denotes the corresponding percentage.}
\vspace{-2mm}
\label{weakly}
\end{figure}

We extend our method to fully supervised methods by replacing manually labeled ground truth with our generated predictions on training set. To be specific, we infer predictions using our trained model on DUTS-Train dataset and adopt CRF for a further refinement.v
It can be seen in Figure {\color{red}\ref{weakly}} that trained with our offered predictions as supervision, BASNet \cite{Qin_2019_CVPR} and ITSD \cite{Zhou_2020_CVPR} achieve $95.6$\% and $97.3$\% of their fully supervised accuracy on F-measure without any pixel-level annotations. 
Additionally, our method also achieves $96.5$\% accuracy of its fully supervised accuracy on F-measure. 
These experiments indicate that our method can serve as an alternative to provide pixel-level supervisions for fully supervised SOD methods while maintaining comparatively high accuracy. This costs only $0.32$\% of pixel-level annotation time and labor.

\vspace{-1mm}

\section{Conclusion}

In this paper, we propose a novel self-calibrated training strategy and introduce a saliency-based dataset with well-matched image-level annotations for WSOD. 
The proposed strategy establishes a mutual calibration loop between pseudo labels and network predictions, which effectively prevents the network from propagating the negative influence of error-prone pseudo labels. 
We also argue that cross-domain inconsistency exists between SOD and existing large-scale classification datasets, and impedes the development of WSOD. 
To offer a better choice for WSOD and encourage more contributions to the community, we introduce a saliency-based classification dataset DUTS-Cls to settle this matter well.
Extensive experiments demonstrate the superiority of our method and effectiveness of our two ideas. 
In addition, our method can serve as an alternative to provide pixel-level labels for fully supervised SOD methods while maintaining comparatively high performance, costing only $0.32$\% of labeling time for pixel-level annotation.

\bibliographystyle{IEEEtran}
\bibliography{egbib}

\end{document}